\documentclass[a4paper]{article}

\usepackage{INTERSPEECH2020}
\usepackage{amsmath}
\usepackage{multirow}
\usepackage{enumerate}
\usepackage{caption}
\usepackage{booktabs}% http://ctan.org/pkg/booktabs
\newcommand{\tabitem}{~~\llap{\textbullet}~~}

\title{Fast and Robust Unsupervised Contextual Biasing for Speech Recognition}
\name{Young Mo Kang, Yingbo Zhou}
\address{
  Salesforce.com, Inc.}
\email{ykang@salesforce.com}

\begin{document}

\maketitle
\begin{abstract}
Automatic speech recognition (ASR) system is becoming a ubiquitous technology. Although its accuracy is closing the gap with that of human level under certain settings, one area that can further improve is to incorporate user-specific information or context to bias its prediction.
A common framework is to dynamically construct a small language model from the provided contextual mini corpus and interpolate its score with the main language model during the decoding process. 

Here we propose an alternative approach that does not entail explicit contextual language model. Instead, we derive the bias score for every word in the system vocabulary from the training corpus.
The method is unique in that 1) it does not require meta-data or class-label annotation for the context or the training corpus. 2) The bias score is proportional to the word's log-probability, thus not only would it bias the provided context, but also robust against irrelevant context (e.g. user mis-specified or in case where it is hard to quantify a tight scope). 3) The bias score for the entire vocabulary is pre-determined during the training stage, thereby eliminating computationally expensive language model construction during inference.

We show significant improvement in recognition accuracy when the relevant context is available.
Additionally, we also demonstrate that the proposed method exhibits high tolerance to false-triggering errors in the presence of irrelevant context.

\end{abstract}


\section{Introduction}

With the recent advances in artificial neural network architectures, ASR performance is rapidly closing its gap with that of human~\cite{asr_performance1, asr_performance2} and widening its applications.
% YZ: please include some citations when you refer the human accuracy performance. e.g. paper from ibm/microsoft on swbd
Today, ASR system is prevalent in our everyday lives, not just limited to generating captions to meetings/videos, but also serving as front-end to numerous downstream tasks such as speech translation, virtual personal assistant, and voice-activated navigation systems. 

Despite its success in general transcription task, typical ASR system suffers from accurately recognizing domain-specific or personalized words and phrases.
This is because the system's language model is trained with finite amount of general data, whose distribution could differ from the target speech context.
For example, names in user's contact list are usually out-of-vocabulary (OOV) and are likely to have very low language model score, thereby making it difficult to accurately predict.
These contextual terms can be personal, such as names in the user's contact~\cite{1326104, 7178957}, current location~\cite{43819, 7846273}, and songs in the playlist~\cite{7846301}; topic-specific, such as medical domain~\cite{medical}; or trending terms~\cite{43816}.
In all these scenarios the contextual information is not static and therefore needs to be dynamically incorporated into the language model during the inference stage.

In~\cite{43819, 7846273, 43816} an explicit contextual language model is dynamically constructed from the provided context during inference. 
This compact n-gram model is interpolated with the larger static general language model and the combined score is used for speech decoding.
Several different interpolation functions and heuristics are investigated to mitigate the problem when data is too scarce to accurately represent the probabilities of n-grams.

In the case when meta-data are available, an entire class of terms can be biased~\cite{1326104, 7178957, 43819, 7472820, token_passing}.
The general idea is to replace every instance of phrases with its class-label to construct a class-based language model~\cite{brown-etal-1992-class}, and dynamically expand the decoding graph of the class-label into class instances provided in the context during inference.
In this framework, class-labels for the provided contextual information as well as matching labels in the train corpus are required, limiting its applications.

In this paper, we present an alternative method for assigning the bias score for each provided context term and demonstrate that the method not only effectively improves accuracy when relevant context is available but also is robust against false-triggering errors from irrelevant terms. Notably, our method differs from previous methods in the following aspects
\begin{itemize}
	\item the method eliminate the need to dynamically construct explicit contextual language model
	\item no meta-data or class-annotations are needed for the context as well as the training corpus
	\item bias score for every vocabulary is pre-computed during the training stage
	\item bias scores are distributed in a way that suppresses false triggering errors.
\end{itemize}

\section{Methodology} \label{methodology}
\subsection{Preliminary}
Let $G$ be the general language model that is constructed from the training corpus, $s_G(\cdot)$ be the score (\emph{i.e.,} log probability) from language model $G$, and let $\mathcal{V}$ be the vocabulary of $G$.
For simplicity, we assume that $G$ is an n-gram language model, but the method can easily be applied to neural language model as well.
We will begin our discussion by limiting the context as a list of $N$ words $\mathcal{B} = (w_1, w_2, \ldots, w_N)$.
In Section~\ref{phrases} we will expand the scenario where the context consists of phrases as well.
The output of the language model fed into the decoder is n-gram score $s(w|H)$ where $w$ is the current candidate word and $H$ is the history. 
In the baseline case where there is no context provided, $s(w|H) = s_G(w|H) = \log(P_G(w|H))$ is the log-probability of the n-gram from the model $G$.

\subsection{Bias Score}
Rather than modeling as an interpolation, we model the contextual bias as a boost score that is added to the base score, similar to~\cite{7846273}
\begin{equation} \label{interpolation}
s(w|H) = s_G(w|H) + s_B(w|H)
\end{equation}
where $s_B(\cdot)$ is the bias score. This boost modeling makes it intuitive to interpret bias qualitatively: positive bias favors the model to predict the given word, negative bias discourages, and zero bias coincides with the original language model score.

%YZ: in general, the description of the method could get expanded a bit more with more details. It would also be helpful to have one or two figures to illustrated 1) the overall flow of the method, 2) the handling of the full matching

\begin{table}[t]
\caption{Overall flow of the proposed method.} \label{flow}
\centering
\begin{tabular}{l}
\toprule
\multicolumn{1}{c}{\textbf{Training Stage}} \\ \midrule
\tabitem Build class-based language model $C$ \\
\tabitem Store $P_C(w|C(w))$ for $ \forall w \in \mathcal{V}$ \\
\tabitem  Discard $C$
\\ \toprule
\multicolumn{1}{c}{\textbf{Inference Stage}} \\ \midrule
\tabitem Evaluate $s_B(w|H)$ and $s(w|H)$ \\
\tabitem Decode with $s(w|H)$\\\bottomrule
\end{tabular}
\end{table}

Table~\ref{flow} describes the overall flow of the proposed method. 
During the training stage, we construct a class-based language model $C$ using an unsupervised method such as~\cite{brown-etal-1992-class, clm2, MARTIN199819} and cluster words in $\mathcal{V}$ into non-overlapping classes based on n-gram statistics.
The n-gram probability from the class-based model $C$ is~\cite{brown-etal-1992-class}
\begin{equation} \label{score}
\begin{split}
&P_C(w_n|w_1 \ldots w_{n-1}) \\
 & = P_C(C(w_n) | C(w_1) \dots C(w_{n-1})) \times P_C(w_n | C(w_n))
\end{split}
\end{equation}
where $C(w_i)$ is the class id for the word $w_i$. 
The first term models class-level n-gram and the second term accounts for the word count in the corpus relative to other words sharing similar n-gram statistics. 
The key idea in this method is to assume $P_G(w|H) \approx P_C(w|H)$ and use Eq.~\eqref{score} to decouple $s_G(w|H)$ into intrinsic n-gram nature and its word count.
Now, biasing the word as if it appears more often in the corpus, and yet retaining its n-gram statistics, is to leave the first term intact and crank up the second term, up to unity (\emph{i.e.,} probability of one).
In other words, $s_B(w) \leq -\log P_C(w|C(w))$ describes a reasonable upper bound of the boost we can apply to the word $w$ without disturbing its inherent n-gram statistics.
We therefore define bias function as follows
\begin{equation} \label{bias_function}
s_B(w|H) = \begin{cases} 
- \lambda \log{P_C(w | C(w))} & w \in \mathcal{B}, w \in \mathcal{V} \\
\alpha & w \in \mathcal{B}, w \notin \mathcal{V} \\
0 & w \notin \mathcal{B}
\end{cases}
\end{equation}
where $\lambda \geq 0$ is a scale factor and $\alpha \geq 0$ is a boost score for OOVs. 
Notice that $s_B$ is independent of the history $H$ and dependent only on the current word $w$.
This property allows us to avoid computationally expensive dynamic language model construction overhead at inference.

Because we perform unsupervised clustering, we cannot guarantee that all words of the same category perceived by humans will fall into the same class. For example, with our train corpus, clustered into 5,000 classes using Brown clustering~\cite{brown-etal-1992-class, Liang05semi-supervisedlearning}, we find that while ``Shanghai'', ``Graz'', ``Hyderabad'', and ``Stockholm'' are all clustered into the same class, ``Robert'' and ``William'' are not. This should not pose a problem because it just implies that there are other words in the train corpus sharing more similar n-gram statistics to ``Robert'' than ``William''.

\subsection{Discussion} \label{discussion}
The bias function described in Eq.~\eqref{bias_function} has four notable benefits. First, the method does not construct an explicit contextual language model from the context provided. If $|\mathcal{B}|$ is large,
construction of the language model results in significant overhead during inference. On the other hand, if $|\mathcal{B}|$ is small, the bias score obtained from the model is not an accurate representation of the probability~\cite{7846273}. Because the system relies on the provided context's statistics to assign a bias score, previous methods~\cite{43819, 7846273, 43816} impose burden on accurate extraction and generation of the context. In contrast, the proposed method uses the provided context only to test the condition $w \in \mathcal{B}$, thereby relieving the burden.

Second, the method does not require external meta-data or class annotations from human. It is usually difficult and costly to find labels for arbitrary words or phrases of given categories. Previous work employing class-level language model~\cite{7178957, 43819, 7472820, token_passing} assumes annotations are available for not only the contextual phrases but also for train corpus. In many applications this is not feasible, limiting the type of context applicable. 
%YK: remove unnecessary statements
%When a new type of context is to be incorporated, the training corpus and the language model $G$ must be first updated in order to incorporate the desired class labels. 
The proposed method, in contrast, relies on unsupervised clustering and therefore does not requires explicit class labels for the contextual phrases or the train corpus, opening the door to a wider range of context, including user-provided context.

Third, the proposed method pre-computes the bias score for every word in the vocabulary during the training phase, so there is minimal overhead during inference. In addition, the bias score is static for a fixed word across different users. On the other hand, previous methods output unstable bias scores  in the sense that they fluctuate across different users as $\mathcal{B}$ statistics changes. Consider, for example, two sets of context: $\mathcal{B}_1 = (\text{``Shanghai''})$ and $\mathcal{B}_2 = (\text{``Shanghai'', ``Graz'', ``Hyderabad'', ``Stockholm''})$. Using the method described in~\cite{7178957} would result in different bias scores even for the same word. Specifically, it would yield $s_{B1}(\text{Shanghai}) > s_{B2}(\text{Shanghai})$. To overcome such effect, \cite{7178957} makes use of a complex heuristic function to scale the bias inversely proportional to $|\mathcal{B}|$. With our method, the bias score for a given context word is always fixed.

Lastly, the proposed method prevents over-biasing by imposing an upper bound on the boost score such that it does not alter its n-gram nature within the corpus. Observe that $P(w | C(w)) \approx 1 $ for a dominant word within its class, yielding $s_B(w) \approx 0$. In other words, if the context word $w$ is already a common word in the train corpus and thus already has a high score $s_G$, the bias score will be close to zero, preventing over-biasing. On the other hand, if $w$ is a rare word, then a large bias will be applied to compensate for its low score $s_G$. For example, with the train corpus used in the experiment, we observe ``trade'' and ``barter'' fall into the same cluster with $s_B(\text{trade}) / \lambda = 0.02$ while $s_B(\text{barter}) / \lambda = 4.241$; this agrees with our intuition that ``trade'' is a much more common term than ``barter''.

One may argue that similar distribution can also be modeled from the unigram probability of $G$ and may propose
\begin{equation}\label{unigram}
    s_B(w) = -\lambda \log P_G(w)
\end{equation}
in place of the first condition in Equation~\eqref{bias_function}. We note that this is a special case of the proposed method when the number of clusters in the class-based language model is set to 1. In this case, the first term in Equation~\eqref{score} would be a constant and the second term precisely equals to the unigram $P_G(w)$, yielding Equation~\eqref{unigram}. As we will see in Section~\ref{num_classes} reducing the number of classes in general improves performance with relevant context but makes it more susceptible to false-triggering errors. On the other end, where the number of clusters is equal to $|\mathcal{V}|$. This essentially translates to $P_G(\cdot) = P_C(\cdot)$.

\subsection{Handling Phrase and Out of Vocabulary Words} \label{phrases}
Let us expand the scenario to include phrases in the given context.
We propose two different scoring schemes for biasing a context phrase.

%YZ: maybe include another column to show the log p(w|C) so that reader will understand where the 0.1 and 0.9 come from?
\begin{table}[t]
\caption{Example language model score for the sentence ``world cup is not a cup'' with the context phrase ``world cup''. Top: expansion scheme and Bottom: OOV scheme.}\label{phrase_example}
\centering
\begin{tabular}{ccccc} \toprule
\textbf{3-gram} & $s_G$ & $\log P(w | C(w))$ & $s_B$ \\ \midrule
\textless{}s\textgreater~ world     & -4.1 & -0.5 & 0.5$\lambda$ & \\
\textless{}s\textgreater~world cup & -1.5 & -0.9 & 0.9$\lambda$ & \\
world cup is                       & -1.6 & -0.1 & 0   \\
cup is not                         & -1.4 & -0.3 & 0   \\
is not a                           & -1.0 & -0.2 & 0   \\
not a cup                          & -3.9 & -0.9 & 0  \\
a cup \textless{}/s\textgreater{}  & -1.1 & - & 0 \\ \bottomrule
\textless{}s\textgreater~\textless{}unk\textgreater     & -8.7 & - & $\alpha$   \\
\textless{}s\textgreater~\textless{}unk\textgreater~is & -2.4 & -0.1 & 0  \\
\textless{}unk\textgreater~is not                         & -1.9 & -0.3 & 0   \\
is not a                           & -1.0 & -0.2 & 0   \\
not a cup                          & -3.9 & -0.9 & 0   \\
a cup \textless{}/s\textgreater{}  & -1.1 & - & 0 \\ \bottomrule
\end{tabular}
\end{table}

In the first scheme, we bias individual words in the phrase only if the entire phrase is an exact match. For example, given a context phrase ``world cup'', we apply the biases $s_B(\text{world})$ and $s_B(\text{cup})$ only when the decoding beam contains the complete phrase ``world cup''.
We refer to this scheme as the \emph{expansion} scheme.
Table~\ref{phrase_example} demonstrates this scheme with an example sentence ``world cup is not a cup''.
Notice that the boost $0.9\lambda$ is applied in the 3-gram ``\textless{}s\textgreater{} world cup'' but not in ``not a cup''.
One way of implementing the exact match condition is to add the phrase into the decoder dictionary as a single token, \emph{i.e.}, ``world-cup'' and to apply word-level biases to this token but not to individual words comprising it, \emph{i.e.,} ``world'' or ``cup''.
This method results in multiple beams with the same transcription, such as ``world-cup'' as a single token and ``world cup'' as two tokens.
In such case, the decoder can keep the highest-scoring beam and discard others.
An alternative method is to construct failure arcs to remove premature bias, as described in~\cite{8682336}.

The second scheme treats every context phrase of two or more words as a single OOV context word.
The context phrase is then boosted with a fixed bias score $\alpha$ as in Eq.~\eqref{bias_function}.
The reasoning behind this scheme is that in many cases a context phrase represents a single entity as a whole and its n-gram statistics within the phrase seen in the train corpus may not accurately represent the given context. 
Rather than relying on its intra-phrase statistics from the train corpus, the scheme simply treats the entire phrase as a single unknown word.
As seen in Table~\ref{phrase_example} this scheme removes intra-phrase n-gram scores during decoding.
We refer to this scheme as the \emph{OOV} scheme.

As for context words not in the dictionary, there is nothing that a language model can do other than assigning the unknown token.
All OOV context words will be biased the same, and it is up to the acoustic score to discern the correct word.

% YZ: I think we could expand this part a bit and add some more details on how to ensure strict phrase matching

% YZ: referring to conditions 1,2,3,4 here make it really hard to remember which is which. I think just write out the described property is better, otherwise, the readers need to keep on checking back and forth to know what is exactly referred.

\section{Experiments} \label{experiments}
% YZ: I think we will need more experiments in addition to currently have listed here: 1) speed performance, 2) number of irrelavant hints, 3) sensitivity to hyper-parameters
% YZ: also describe in detail how the hints are constructed
\subsection{Setup} \label{setup}

%\begin{table}[t]
%\caption{Example utterances on the test %set.}\label{categories}
%\centering
%\begin{tabular}{cc} \toprule
%\textbf{Category} & \textbf{Example} \\ \midrule
%Address & five oh two emerson street palo alto \\
%Assist & remember to call taylor swift \\
%Company & salesforce dot com incorporated \\
%Earnings & we had an amazing quarter \\
%Name & einstein e i n s t e i n \\
%Numbers & one three five eight zero six \\
%\bottomrule
%\end{tabular}
%\end{table}

Our ASR system consists of character-level connectionist temporal classification~\cite{ctc, policy_learning} acoustic model and 4-gram language model. We limit the system vocabulary by extracting the top 400k frequent words from the train corpus. 
% YZ: please also cite our own paper here "Improving End-to-End Speech Recognition with Policy Learning"
We create a class-based language model having 500 classes from the train corpus via Brown clustering method~\cite{brown-etal-1992-class, Liang05semi-supervisedlearning}.
We first evaluate the system on our internal test set consisting of 10,643 utterances of various categories. 
From the prediction result, we divide these utterances into two sets: \emph{with-error} set of 7064 utterances having at least one error, and (b) \emph{without-error} set of 3579 utterances matching exactly with the ground truth.
The baseline word error rate (WER) of the \emph{with-error} set is 16.65\%.

For each utterance in the \emph{with-error} set, we create \emph{relevant} context by extracting words and phrases from the target that are missing from the prediction.
We limit the maximum length of context phrase to be three-words to emulate a more realistic scenario.
The average number of context phrases per utterance is 1.6 and the average number of words per phrase is 1.3.
We refer to the relevant phrases as \emph{oracle} context.

We also create \emph{irrelevant} context by randomly selecting 10,000 phrases of length one-, two-, and three-words from the internal test set such that the phrases are not part of the \emph{with-error} set target transcriptions.
Because both the relevant and irrelevant context are derived from the same pool, it presents a harsher condition to the system when testing false-triggering errors.
We refer to the irrelevant phrases as \emph{distractors}.

\subsection{Oracle Test}

\begin{table}[t]
\caption{Oracle test results on the with-error set with phrase expansion (top) and OOV (bottom) schemes.}\label{oracle_test}
\centering
\begin{tabular}{cc|cccc}
\toprule
\multicolumn{2}{c|}{\multirow{2}{*}{\textbf{WER(\%)}}} & \multicolumn{4}{c}{$\lambda$} \\
\multicolumn{2}{c|}{} & 0 & 0.5 & 1 & 1.5 \\ \midrule
\multirow{3}{*}{$\alpha$} & 0 & 15.41 & 13.42 & 12.20 & 11.54 \\
                              & 2.5 & 14.63 & 12.64 & 11.39 & 10.74 \\
                              & 5 & 14.01 & 12.01 & 10.77 & 10.13 \\ \midrule
\multirow{3}{*}{$\alpha$} & 0 & 14.67 & 13.34 & 12.66 & 12.39 \\
                              & 2.5 & 13.54 & 12.22 & 11.53 & 11.26 \\
                              & 5 & 12.38 & 11.07 & 10.37 & 10.12    \\ \bottomrule
\end{tabular}
\end{table}

We first experiment to see how much the system performance improves with the relevant context.
In practice we cannot deduce the \emph{oracle} context beforehand, but this experiment is useful in that it provides upper bound on how much the system can improve with the relevant context.
We compare the system WER on the \emph{with-error} set with the relevant context for different values of hyperparameters $\lambda$ and $\alpha$ in Table~\ref{oracle_test}.
Notice minor performance improvement with zero bias applied in $\lambda = \alpha = 0$ case, as OOV words are included in the decoding dictionary.
The result shows that the \emph{expansion} scheme is more sensitive to $\lambda$ while \emph{OOV} scheme is more sensitive to $\alpha$.
This is because the bias score of a phrase of length two or more words is affected by $\lambda$ in the former case but by $\alpha$ in the latter case.
Compared to the baseline WER of 16.65\%, providing relevant context helps the system improve performance by close to 6\% points at $\lambda = 1$ and $\alpha = 5$, and even further with larger $\lambda$ and $\alpha$.
In practice we cannot arbitrarily increase  $\lambda$ and $\alpha$ because of over-biasing effect when irrelevant context is provided.
This leads us to the distractors test to assess its tolerance to false-triggering errors.

\subsection{Distractors Test}

\begin{figure}[t]
\centering
  \includegraphics[width=0.45\textwidth]{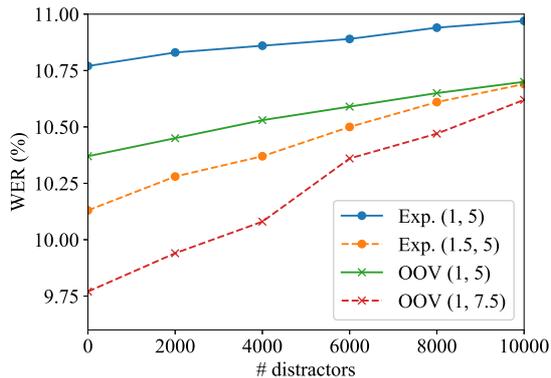}
  \caption{Distractors test WER results on with-error set with hyperparameter $(\lambda, \alpha)$.}
  \label{distractors_test}
\end{figure}

In this test, we evaluate system performance with different numbers of distractors (a) \emph{with} and (b) \emph{without} the relevant context.
Figure~\ref{distractors_test} shows WER result for the \emph{with-error} set. 
We observe that the introduction of distractors degrades the ASR performance in a roughly linear fashion with respect to the number of distractors.
The system is robust against false-triggering in that even with 10,000 distractors the WER increases only by 0.20\% and 0.33\% point for the \emph{expansion} and \emph{OOV} schemes, respectively.

In practice one should optimize hyperparameters $\lambda$ and $\alpha$ for a specific operating region of interest.
For example with the phrase \emph{OOV} scheme, we would choose $\alpha = 7.5$ over $\alpha = 5$ if we are to provide up to 10,000 context phrases, because the WER with $\alpha = 7.5$ is consistently lower than that with $\alpha = 5$.
On the other hand, if we are to provide a larger number of context, say up to 100,000 phrases, then we would choose $\alpha = 5$ because its WER is likely to be lower for $>$10,000 context phrases, as the crossing point is roughly at 10,000 distractors.

We repeat the experiment for (b) and observe that the WER degradation is consistently lower than that of (a), and thus not reported here.
We also repeat the test on \emph{without-error} set; the WER degradation with 10,000 distractors is 0.48\% and 0.37\% point for \emph{expansion} and \emph{OOV} schemes, respectively for $\lambda = 1$ and $\alpha = 5$.

\subsection{Number of Classes} \label{num_classes}

\begin{table}[t]
\caption{WER results on different number of classes with phrase expansion (left) and OOV (right) schemes.}\label{classes_test}
\centering
\begin{tabular}{cc|cccc}
\toprule
\multicolumn{2}{c|}{\multirow{2}{*}{\textbf{WER (\%)}}} & \multicolumn{2}{c|}{\# Distractors} &  \multicolumn{2}{c}{\# Distractors}               \\
\multicolumn{2}{c|}{} & 0 & \multicolumn{1}{c|}{10000} & 0 & 10000 \\ \midrule
\multirow{3}{*}{\# Classes} & 50 & 9.55 & \multicolumn{1}{c|}{10.40} & 9.82 & 10.26 \\
& 500 & 10.77 & \multicolumn{1}{c|}{10.97} & 10.37 & 10.70 \\
 & 5000 & 12.40 & \multicolumn{1}{c|}{12.43} & 11.16 & 11.44 \\ \bottomrule
\end{tabular}
\end{table}

Number of classes of $C$ is another hyperparameter.
We repeat the distractors test using different numbers of classes. 
Table~\ref{classes_test} compares the WER results on the \emph{with-error} set with 50, 500, and 5,000 classes at $\lambda = 1$ and $\alpha = 5$.
In general, we observe better performance with fewer classes but at the expense of higher false-triggering error. 
This is because fewer classes will translate to, on average, lower class-conditional unigram (the second term in Equation~\eqref{score}), hence resulting in higher bias score for each word.
Likewise, we observe that the system becomes less sensitive to its hyperparameters $\lambda$ and $\alpha$ with larger number of classes.
Notably the \emph{expansion} method exhibits mere 0.03\% point increase in WER using 5,000 classes with 10,000 distractors, achieving remarkable tolerance against false-triggering errors, while still improving by more than 4\% points with the relevant context.
%YZ: could add a small relevant work session here, no need to be very detailed
%YK: the page limit is 4 w/o references; we don't have space

\subsection{Adversarial Attack}

One way to intentionally over-bias and corrupt the system is to provide it with context consisting of common words, such as ``the'', ``to'', ``of'', etc that already have high n-gram score $s_G(w|\cdot)$ but do not represent the target domain. 
In practice this attack can be mitigated by filtering out the context of common words.
Here we evaluate the method's intrinsic tolerance to such an \emph{adversarial attack} without any external counter-measure.
We randomly select 10,000 out of top 100,000 most common words that have the highest unigram score $s_G(w)$.
We run the system with these common words as context on the \emph{with-error} set at $\lambda = 1$, $\alpha = 5$. The resultant WER is measured to be 18.68\%, 17.38\%, and 16.81\% for 50, 500, and 5,000 classes, respectively.
As with the distractors test, using a larger number of classes reduces over-bias effect with the attack.

\subsection{Run Time Overhead}
% baseline 380ms
% with 10000 context phrases 704ms
% with non-OOV 10000 context words 470ms
% kenlm 3-gram build 1288ms

Here we report rough decoding time overhead during inference using the proposed method.
The decoding time during offline inference of a 10-seconds-long utterance is measured to be 380 ms without context and 704 ms with 10,000 context phrases.
The increase in run time with the context is largely contributed by adding OOV context into the decoding dictionary and is not directly contributed by this method.
To isolate this overhead, we repeat the experiment with 10,000 context words selected from the system vocabulary, reducing the decoding time to 470 ms.
As a reference, user execution time of building a 3-gram language model from the same 10,000 context phrases using KenLM~\cite{kenlm} is measured to be 1,288 ms. This excludes time spent blocked or in system calls.
All experiments are run on a Docker container~\cite{docker} constrained with available processors to a single core, emulating a single-threaded execution.
The proposed method avoids construction of the contextual language model, which is significant bottleneck during inference.

\section{Conclusions}

We present a fast and robust method for assigning bias scores for provided context in an ASR system. The method is unique in that it does not require explicit construction of contextual language model and instead derives bias scores from the train corpus, eliminating overhead during the inference. The method also employs an unsupervised clustering method, thereby widening its applications beyond context with known meta-data or class-labels. Lastly, the method exhibits high tolerance to false-triggering errors.

\section{Acknowledgement}
We thank Sreya Basuroy for helpful feedback and suggestions.

\bibliographystyle{IEEEtran}

\bibliography{mybib}

\end{document}